\documentclass[10pt, a4paper]{article}

\usepackage[final]{lrec2026} % this is the new style
\usepackage{booktabs}
\usepackage{paralist}
\usepackage{verbatim}
\usepackage{multirow}
\usepackage{amsmath}

\title{FineDialFact: A Benchmark for Fine-Grained Dialogue Fact Verification}

\name{Xiangyan Chen\textsuperscript{\rm 1}, Yufeng Li\textsuperscript{\rm 1}, Yujian Gan\textsuperscript{\rm 2}, Arkaitz Zubiaga\textsuperscript{\rm 1}, Matthew Purver\textsuperscript{\rm 1, 3}} 

\address{\textsuperscript{\rm 1}Queen Mary University of London, UK \\ \textsuperscript{\rm 2} Queen's University Belfast, UK\\ \textsuperscript{\rm 3} Institut Jožef Stefan, Slovenia \\
         \{xiangyan.chen, yufeng.li, a.zubiaga, m.purver\}@qmul.ac.uk, y.gan@qub.ac.uk}

\abstract{
Large language models are known to produce hallucinations --- factually incorrect or fabricated information --- which poses significant challenges for many natural language processing applications, such as dialogue systems. As a result, detecting hallucinations has become a critical area of research. Current approaches to hallucination detection in dialogue systems primarily focus on verifying the factual consistency of generated responses. However, these responses often contain a mix of accurate, inaccurate or non-verifiable facts, making the use of a single factual label overly simplistic and coarse-grained. In this paper, we introduce a benchmark, FineDialFact, for fine-grained dialogue fact verification, which involves verifying atomic facts extracted from dialogue responses. To support this, we construct a dataset based on publicly available dialogue datasets and evaluate it using various baseline methods. Experimental results demonstrate that methods incorporating Chain-of-Thought reasoning can enhance performance in dialogue fact verification. Despite this, the best F1-score achieved on the HybriDialogue, an open-domain dialogue dataset, is only 0.74, indicating that the benchmark remains a challenging task for future research. We release our dataset and code at \url{https://github.com/XiangyanChen/FineDialFact}. 
\\ \newline \Keywords{large language models, fine-grained dialogue fact verification, dialogue hallucination detection} }

\begin{document}

\maketitleabstract
\section{Introduction}
In recent years, large language models (LLMs) have demonstrated impressive capabilities across a wide range of tasks \cite{zhao2023survey}. However, one persistent challenge is hallucination --- the generation of factually incorrect or misleading content. This issue is particularly concerning in dialogue systems, where hallucinated responses can mislead users and potentially pose risks to social trust and stability \cite{ji2023survey}.

\begin{figure}[t]
\centering
  \includegraphics[width=0.9\linewidth]{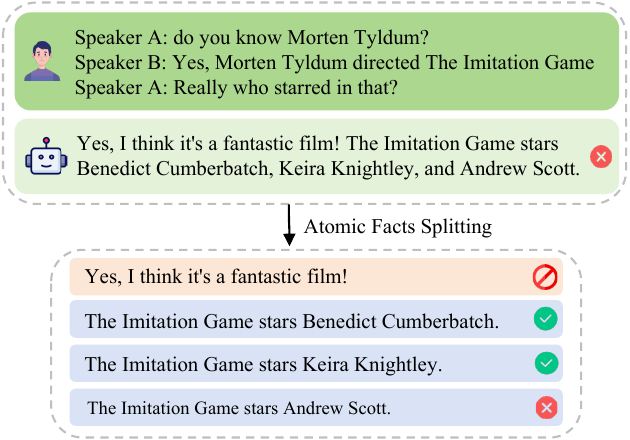}
  \caption {An example of the response-level based dialogue fact verification and fine-grained one. The difference between them is that the latter is based on the atomic facts. The prohibition symbol means it is not a verifiable factual claim.}
  \label{figure:example}
\end{figure}

Previous approaches to hallucination detection in dialogue systems mainly rely on human evaluation \cite{ni2023multi, li2022eliciting, shuster2021retrieval, yu2022xdai}, which is time-consuming and labour-intensive. Recently, automatic methods have been proposed, including uncertainty estimation \cite{farquhar2024detecting} and fact verification \cite{chen2024diahalu}. However, uncertainty estimation often fails when the model is overconfident in hallucinated content. Alternatively, our work focuses on direct fact verification. Existing fact verification methods for dialogue systems \cite{chen2024diahalu, gupta2021dialfact} verify responses using external knowledge and dialogue context, and output one of three labels: \emph{Supports}, \emph{Refutes}, or \emph{Not Enough Information}. Yet, these methods operate only at the response level, ignoring that a single response may contain all factual, hallucinated, and non-verifiable information. As shown in Figure~\ref{figure:example}, labelling the entire response as incorrect is overly coarse since it also contains accurate facts.

To address the above limitation, we systematically study fine-grained fact verification for dialogue systems and offer a benchmark, named \emph{FineDialFact}. We process the dialogue response into small, verifiable sentences, called atomic facts. Each atomic fact expresses exactly one proposition that can be judged and then is verified independently using external knowledge and LLMs. As no existing datasets are available, we construct a dataset by extending two public dialogue datasets, OpenDialKG and HybriDialogue, and report their inter-annotator agreement using Cohen's Kappa \cite{cohen1960coefficient}. To evaluate the dataset, we provide a set of metrics: accuracy, precision, recall, F1-score, Geometric Mean (G-mean), and Cohen's Kappa, which measure performance from different perspectives. 

In addition, we evaluate a series of Chain-of-Thought (CoT) based approaches using different LLMs, such as Llama3 \cite{dubey2024llama}, Deepseek-R1 \cite{guo2025deepseek}, QwQ \cite{team2025qwq}, and GPT \cite{hurst2024gpt}. These CoT approaches include zero-shot CoT, few-shot CoT, and CoT distillation. For zero-shot CoT, we simply add a reasoning prompt. For few-shot CoT, we manually annotate a set of samples and use GPT-4o to generate corresponding reasoning steps, then retrieve the top-N most relevant samples as demonstrations. For CoT distillation, we use GPT-4o to annotate data and generate CoT reasoning processes, which are then used to fine-tune smaller language models. The experimental results show that CoT series-based methods are able to improve the performance of LLMs significantly. However, fact verification on the manually annotated HybriDialogue dataset remains challenging, with the best F1-score of 0.74 achieved by Llama-3.3-70B.

The contributions can be listed as follows:

\begin{compactenum}

\item We delve into fine-grained fact verification for dialogue systems by developing a novel benchmark named FineDialFact. To the best of our knowledge, this is the first systematic research on the fine-grained factuality evaluation of dialogue systems.

\item We provide a newly constructed dataset, including manually and automatically annotated data, to evaluate the fine-grained dialogue factuality, laying a foundation for further research in this area.

\item We evaluate several baselines, including CoT series approaches. The experimental results show that the human-annotated HybriDialogue dataset is more challenging, and the highest score, achieved by Llama-3.3-70B, is only 0.74, opening up new challenges for future research.

\end{compactenum}

\section{Related Work}
\subsection{Dialogue Hallucination Detection}
Hallucination in LLMs has attracted increasing attention in open-domain dialogue systems. While current hallucination detection approaches often rely on human evaluation \cite{ni2023multi, yu2022xdai}, this method is time-consuming and labour-intensive, highlighting the need for effective automatic evaluation methods. \citet{chen2024diahalu} and \citet{gupta2021dialfact} addressed a critical gap with a dialogue-level fact-verification and hallucination detection benchmark that extends beyond factuality. 

Despite these advances, current methods still struggle with dialogue responses that mix correct and incorrect information.

\subsection{Fine-grained Detection}
Inspired by the challenges observed in dialogue hallucination detection, we further review fine-grained detection techniques developed in general domains.  Aspect-based sentiment analysis acknowledges the possibility of having both positive and negative sentiments in the same sentence \cite{tan2019recognizing}. \citet{song2024finesure} and \citet{wan2024acueval} introduced fine-grained techniques for detecting hallucinations in text summarisation, allowing for more accurate identification of factual errors. \citet{min2023factscore} proposed a fine-grained fact scoring method to evaluate factual accuracy in long-form text generation, although its use has so far been limited to bio-generation. Similarly, \citet{mitra2024factlens} provided a fact verification benchmark aimed at splitting claims into sub-claims and analysed the importance of the quality of sub-claims.

Nevertheless, the above studies have not addressed the unique challenges of the dialogue domain, where fact verification must account for the evolving conversational context, making the task more complex. To this end, we introduce a fine-grained approach for dialogue fact verification.

\subsection{Chain of Thought}
Since the rise of LLMs, there has been growing interest in applying them to NLP tasks. The CoT approach \citep{wei2022chain} improves performance on complex tasks by introducing intermediate reasoning steps. To reduce the need for hand-crafted few-shot examples, \citet{zhang2022automatic} proposed automatically collecting examples via clustering. \citet{kojima2022large} showed that zero-shot CoT can also work well by prompting with ``Let’s think step by step.'' CoT has also been applied in training, such as in CoT-based knowledge distillation \citep{li2023symbolic}, where transferring reasoning to smaller models boosts performance.

\section{The FineDialFact Benchmark}
Previous works \cite{gupta2021dialfact, chen2024diahalu} on dialogue fact verification focus solely on whether the response is factually correct or has insufficient information to make a judgment. However, a response may contain factually correct and incorrect facts, as well as non-verifiable factual claims, and only verifying the response is coarse-grained. 

To detect hallucinations in dialogue at a fine-grained level, we aim to verify each verifiable atomic fact extracted from the response. As there are no existing dialogue datasets containing atomic facts, we build one extending public dialogue datasets: (1) we generate dialogue response by LLMs as sampling hallucinated examples, see Section~\ref{Sec:Dialogue response geneation} for details; (2) we split the dialogue response into atomic facts based on few-shot learning, as described in Section~\ref{Sec:Atomic fact splitting}; (3) we describe retrieving knowledge (Section~\ref{Sec:Knowledge Retriever}), manual (Section~\ref{Sec:Dataset Annotation}) and automated data annotation (Section~\ref{sec:automated_annotation}), and evaluation metrics (Section~\ref{Sec:metrics}).

\subsection{Hallucinated Data Sampling}
\label{Sec:Dialogue response geneation}
To construct our dataset, we select two public knowledge-grounded datasets: OpenDialKG and HybriDialogue. OpenDialKG \cite{moon2019opendialkg} includes a recommendation component focused on movies and books, along with a chit-chat component centred around sports and music. HybriDialogue \cite{nakamura2022hybridialogue} is an open-domain dialogue dataset designed for information-seeking conversations.  However, it only offers dialogue references with factually correct facts, and for fine-grained fact verification, the hallucinated samples are needed. Instead of prompting LLMs to produce hallucinated content, we guide them to generate responses based on given dialogues, ensuring the samples are consistent with the dialogue style. 

We adopt various LLMs to generate dialogue responses, such as Llama-3.1-8B-Instruct, Flan-T5-XXL \cite{chung2024scaling}, to ensure inclusiveness. %The detailed prompt is listed in the Appendix~\ref{Appendix:prompts}.

\subsection{Atomic Fact Splitting}
\label{Sec:Atomic fact splitting}
%\citet{min2023factscore} proposed that fact scores should be computed based on atomic facts, the smallest units of factual information, rather than long texts.
We define atomic facts as minimal, independent propositions in line with \citet{min2023factscore}, and we verify each unit separately with external evidence and an LLM-based judge.

In this work, we follow \citet{min2023factscore} setup to decompose responses into multiple atomic facts using LLMs. The splitting process relies on few-shot learning, with two examples retrieved through BM25 \cite{robertson2009probabilistic}. The atomic splitting is construction-agnostic and is guided by the few-shot demonstrations.  The dialogue scenario contains some non-verifiable atomic claims, such as opinions, which are not considered to be verified further. For reproducibility, the open-source model Llama-3-70B-Instruct is used in the atomic fact splitting.

We randomly selected 100 samples from the HybriDialogue dataset to assess the quality of atomic fact splitting. Atomic facts were rated as Good (1), Acceptable (0.5), or Bad (0) based on accuracy, completeness, and clarity. The raw agreement between annotators is 0.85, with a Cohen’s kappa of 0.562, indicating moderate agreement. The average score is 0.883, showing high-quality atomic fact splitting. %The prompt for atomic fact splitting is detailed in Appendix~\ref{Appendix:prompts}.

\subsection{Knowledge Retriever}
\label{Sec:Knowledge Retriever}
Due to the ubiquity of LLM hallucinations, the internal knowledge is unreliable. Therefore, the models rely on external knowledge to verify. We adopt sophisticated \emph{Contriever-MS MARCO} \cite{izacard2021unsupervised} as our retriever, which is designed by contrastive learning, achieving good performance on document retrieval. 

We use Wikipedia as our knowledge source, dividing each article into fixed-length passages, since full articles are often too long for LLMs to process.

\begin{table}[t]
\centering
\resizebox{0.48\textwidth}{!}{%
\setlength{\tabcolsep}{1pt}
\begin{tabular}{l|cc|cc|cc|cc}
\midrule
\multirow{4}{*}{\textbf{Task}} 
& \multicolumn{4}{c|}{\textbf{Humans}} 
& \multicolumn{4}{c}{\textbf{GPT-4o}} \\
\cmidrule(lr){2-5} \cmidrule(lr){6-9}
& \multicolumn{2}{c|}{\textbf{HD}} & \multicolumn{2}{c|}{\textbf{ODKG}} 
& \multicolumn{2}{c|}{\textbf{HD}} & \multicolumn{2}{c}{\textbf{ODKG}} \\
\cmidrule(lr){2-3} \cmidrule(lr){4-5} \cmidrule(lr){6-7} \cmidrule(lr){8-9}
& \textbf{Agr.} & \textbf{Kappa} & \textbf{Agr.} & \textbf{Kappa} & \textbf{Agr.} & \textbf{Kappa} & \textbf{Agr.} & \textbf{Kappa} \\
\midrule
Factual Claim & 0.865 & 0.661 & 0.865 & 0.692 & 0.864 & 0.657 & 0.848 & 0.687 \\
Factual Label & 0.722 & 0.546 & 0.735 & 0.533 & 0.754 & 0.594 & 0.840 & 0.717 \\
\midrule
\end{tabular}}
\caption{Comparison of agreement for two annotation tasks: The left panel shows human inter-annotator agreement for HybriDialogue (HD) and OpenDialKG (ODKG), and the right panel shows GPT‑4o with gold labels. Metrics include raw agreement (Agr.) and Cohen’s kappa.}
\label{tab:agreement_combined}
\end{table}

\begin{table}[t]
\centering
\resizebox{0.45\textwidth}{!}{
\setlength{\tabcolsep}{3pt}
\begin{tabular}{l|cc|cc}
\midrule
\multirow{2}{*}{\textbf{Task}} 
& \multicolumn{2}{c|}{\textbf{Humans}} 
& \multicolumn{2}{c}{\textbf{GPT-4o}} \\
\cmidrule(lr){2-3} \cmidrule(lr){4-5}
 & \textbf{HD} & \textbf{ODKG} & \textbf{HD} & \textbf{ODKG} \\
\midrule
Evidence Selection & 0.577 & 0.556 & 0.642 & 0.590 \\
\midrule
\end{tabular}}
\caption{Annotation overlap (Jaccard Similarity) results for the evidence selection task. The left shows inter-annotator overlap, and the right shows overlap between ground truth and GPT-4o.}
\label{tab:agreement_evidence_combined}
\end{table}

\subsection{Manual Data Annotation}
\label{Sec:Dataset Annotation}
%Since some decomposed atomic claims are non-factual, e.g opinions, which should not be considered in dialogue fact verification. 

After collecting dialogue responses generated by LLMs, we randomly mix them with reference responses from public dialogue datasets. We then extract atomic facts from these samples and retrieve the top N relevant knowledge passages from Wikipedia using the Contriever-MS MARCO retriever. %The retrieval is based on semantic matching between Wikipedia texts and the combination of atomic facts and dialogue history.
Retrieval is performed via semantic matching between Wikipedia passage embeddings and a query embedding, using cosine similarity as the similarity metric. The query is constructed from the atomic facts and the dialogue history.

We then ask two annotators to follow a three-step annotation process: first, they assess the verifiability of each factual claim; second, they select the most relevant Wikipedia passages as evidence for each verifiable atomic fact; third, they verify each atomic fact against the selected evidence and dialogue history, assigning one of three labels --- \emph{Supports}, \emph{Refutes}, or \emph{Not Enough Information}. All annotators perform these steps independently. Since the datasets are in English, only annotators with good English proficiency are selected. None of them have prior experience with AI or LLMs, nor familiarity with hallucination phenomena, which helps reduce potential biases in the annotations.

After annotation, we assessed the similarity of the knowledge source by Jaccard Similarity (JS) \cite{jaccard1901etude}. We measured the agreement using Cohen's Kappa \cite{cohen1960coefficient}, which considers chance agreement and is widely used in NLP annotation tasks.
When there was a disagreement, we asked a third annotator to choose a factual label by majority vote among the previous annotators. 

We sampled 500 atomic claims each from HybriDialogue and OpenDialKG (1,000 total). As shown in Table~\ref{tab:agreement_combined}, factual-claim annotation attains an agreement of 0.865 on both datasets with substantial consistency (Cohen’s kappa > 0.6). Fact labelling yields agreements of 0.722 and 0.735 with moderate consistency (kappa > 0.5). Human knowledge-source choices have JS of about 0.6 on both datasets, indicating moderate overlap. After filtering non-factual claims, the distribution of factual labels is reported in Table~\ref{tab:fdf_merged}.

\begin{table}[t]
\centering
\scriptsize
\begin{tabular}{l|cc|c|cc|c}
\toprule
\multirow{3}{*}{\textbf{Label}} & \multicolumn{3}{c|}{\textbf{Humans}} & \multicolumn{3}{c}{\textbf{GPT-4o}} \\
 \cmidrule(lr){2-4} \cmidrule(lr){5-7}
 & \textbf{HD} & \textbf{ODKG} & \textbf{FDF} & \textbf{HD} & \textbf{ODKG} & \textbf{FDF} \\
\midrule
Supports        & 181 & 200 & 381 & 1,833 & 2,153 & 3,986 \\
Refutes         & 55  & 42  & 97  & 249   & 201   & 450 \\
NEI & 134 & 83  & 217 & 1,896 & 1,723 & 3,619 \\
\midrule
\textbf{Total}  & 370 & 325 & 695 & 3,978 & 4,077 & 8,055 \\
\bottomrule
\end{tabular}
\caption{Comparison of FineDialFact (FDF) label distributions in human and automated datasets. NEI stands for Not Enough Information. HD and ODKG refer to the datasets HyriDialogue and OpenDialKG, respectively.}
\label{tab:fdf_merged}
\end{table}

\subsection{Automated Data Annotation}
\label{sec:automated_annotation}
As an efficient approach to increase dataset size, we adopt GPT-4o for automated data annotation following a three-step process: detecting verifiable claims, evidence selection, and fact verification.

For identifying verifiable factual claims and selecting evidence, we use zero-shot prompting, while few-shot learning is applied for fact verification, as detailed in Section~\ref{Sec:Few-shot CoT Prompting}. We first assess the LLM’s performance on these three tasks (see Table~\ref{tab:agreement_combined} and~\ref{tab:agreement_evidence_combined}); the results show that GPT-4o achieves a Cohen’s kappa and JS of approximately 0.6, indicating substantial agreement with the ground truth and demonstrating its reliability.

We automatically annotated the atomic facts in 500 dialogues from each dataset. Table~\ref{tab:fdf_merged} presents the distribution of the atomic facts. We use GPT-4o to identify the verifiable factual claims and select evidence per dialogue response. These annotations allow us to analyse the relationship between response-level and atomic-level fact verification.

\begin{comment}
\setlength{\tabcolsep}{4pt}
\begin{table}[ht]
\centering
\begin{tabular}{l|cc|c}
\toprule
\textbf{Label} & \textbf{HD} & \textbf{OKG} & \textbf{FDF (total)} \\
\midrule
Supports        & 1,833 & 2,153 & 3,986 \\
Refutes         & 249  & 201 & 450 \\
Not Enough Info & 1,896 & 1,723 & 3619 \\
\midrule
\textbf{Tota\textbf{}l}  & 3,978 & 4,077 & 8055 \\
\bottomrule
\end{tabular}
\caption{Distribution for factual labels in FineDialFact (FDF) and by source: HybriDialogue (HD) and OpendialKG (OKG).}
\label{tab:factuallabel_stats1}
\end{table}
\end{comment}

\begin{figure*}[ht]
\centering
  \includegraphics[width=1\linewidth]{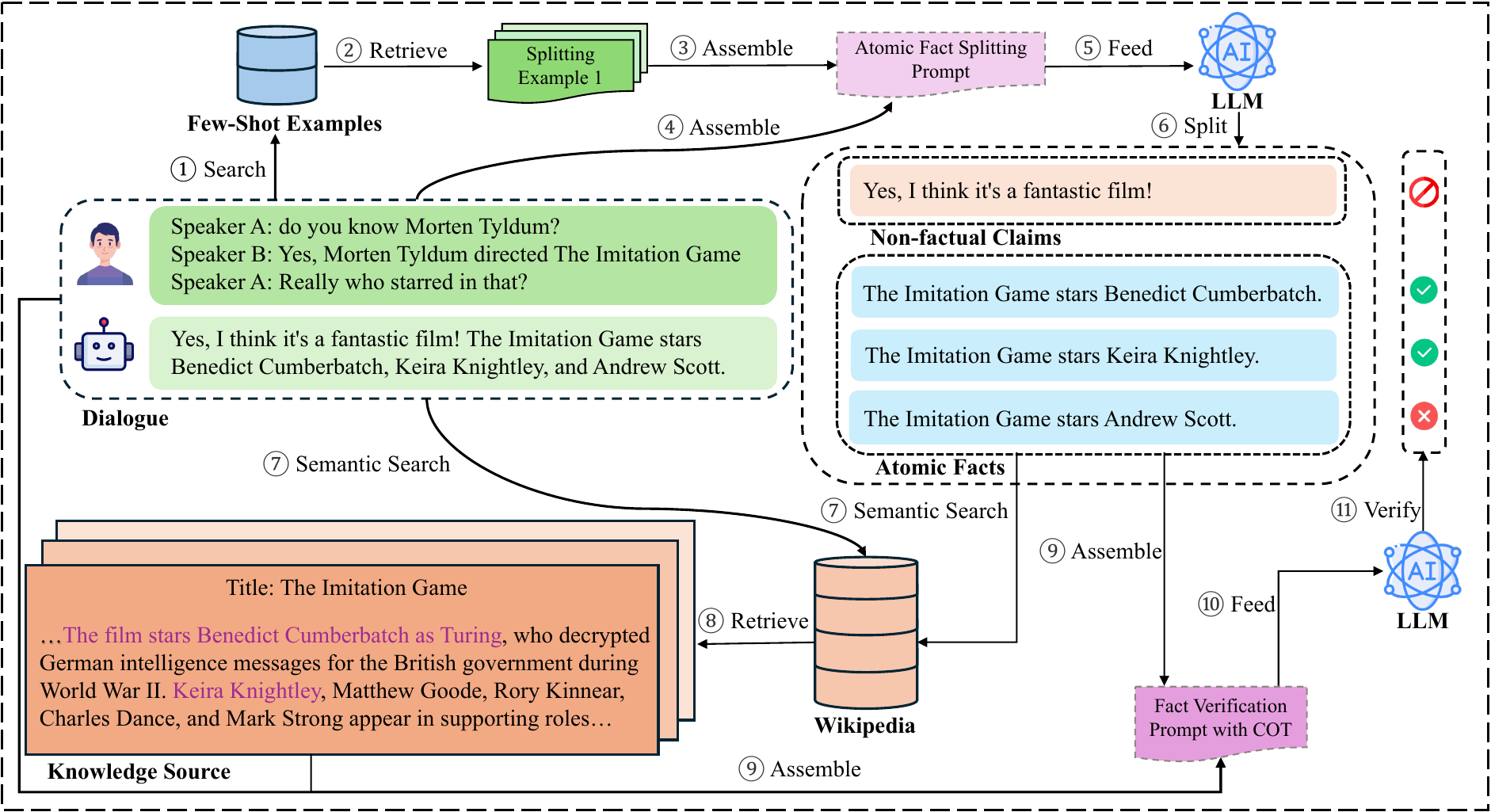}
  \caption {The framework for fine-grained dialogue fact verification. Starting from a dialogue, an LLM splits the response into several atomic facts using few-shot prompting. For each atomic fact, relevant evidence passages are retrieved from Wikipedia via semantic search. An LLM then verifies each atomic fact against the retrieved evidence and outputs the factual label.}%The knowledge source is retrieved from the knowledge database based on the semantic matching of atomic facts and used for precise fact verification.}
  \label{figure:framework} 
\end{figure*}

\subsection{Evaluation Metrics}
\label{Sec:metrics}
We use classification metrics to validate the performance of dialogue fact verification, including accuracy, precision, recall and F1-score. Accuracy reflects the overall performance of a classifier, but it may be misleading when dealing with imbalanced data. The F1-score and Geometric Mean (G-Mean)  can more realistically reflect performance for imbalanced data. The G-Mean is calculated based on the recall scores of different classes. In addition, Cohen's kappa is employed to assess the model-human agreement between the classifier and annotator, thereby reflecting agreement beyond chance.

%In addition, we use raw agreement to measure inter-annotator agreement. However, since raw agreement does not account for chance agreement, we also adopt Cohen's Kappa \cite{cohen1960coefficient} to evaluate inter-annotator reliability. Furthermore, Cohen's Kappa is employed to assess the model-human agreement between the classifier and human annotators, thereby reflecting agreement beyond chance.

\section{Fine-grained Dialogue Fact Verification}
We introduce a framework for fine-grained fact verification in dialogue systems, as illustrated in Figure 1. Building on this framework, we propose CoT baselines to evaluate the datasets, including zero-shot CoT (Section~\ref{Sec:Zero-Shot Chain of Thought}), few-shot CoT prompting (Section~\ref{Sec:Few-shot CoT Prompting}), and CoT distillation (Section~\ref{Sec:Reasoning Distillation}).

\subsection{Task Definition}
We define our task as fine-grained dialogue fact verification. A dialogue is represented as \( \mathcal{U} = \{\text{U}_1, \text{U}_2, ..., \text{U}_m\} \), where \( m \) denotes the number of dialogue turns. The goal is to verify the factual accuracy of the last utterance \( \text{U}_m \). This last utterance is decomposed into a set of atomic facts \( \mathcal{A} = \{a_1, a_2, ..., a_n\} \), where \( n \) is the total number of atomic facts. To verify these facts, relevant knowledge is retrieved in the form of passages \( \mathcal{T} = \{t_1, t_2, ..., t_k\} \), with \( k \) indicating the number of retrieved passages. For few-shot learning, we retrieve examples defined as $\mathcal{E} = \{e_1, e_2, ..., e_l\}$, where \( l \) is the number of examples. Each atomic fact is then classified into one of three labels: \emph{Supports}, \emph{Refutes}, and \emph{Not Enough Information}, based on the retrieved knowledge.

\subsection{Zero-Shot Chain-of-Thought}
\label{Sec:Zero-Shot Chain of Thought}
Different from traditional fact verification, dialogue history containing a large number of pronoun references should also be considered when verifying facts in dialogue settings, making the task more complex. CoT \cite{wei2022chain} is a prompting strategy for solving complex tasks. The original CoT requires few reasoning examples. But \citet{kojima2022large} proposed a zero-shot CoT, showing that adding ``let's think step by step'' to the prompt can remarkably improve LLM performance. 

CoT has been shown to lead to competitive performance in dialogue fact verification. To verify dialogue facts, we ask the LLM if an atomic fact $a_i$ is factually correct against external knowledge $\mathcal{T}$ and dialogue history $\mathcal{U}_{1:m-1}$. And we simply add ``think step by step'' into the fact verification prompt. The formula is listed as follows:

\begin{equation}
(o_i^{\text{reason}}, o_i^{\text{label}}) = \mathcal{M}_\theta(p_{\text{fact}}, a_i, \mathcal{U}_{1:m-1}, \mathcal{T}),
\end{equation}
where $o_i^\text{reason}$ and $ o_i^\text{label}$ denote outputs of the reasoning process and factual label, including factual label and reasoning steps. $\mathcal M$ is the LLM for fact verification. $p_\text{fact}$ is the prompt template for verifying facts, %described in Appendix~\ref{Appendix:prompts}.

\subsection{Few-shot Chain-of-Thought Prompting}
\label{Sec:Few-shot CoT Prompting}
Few-shot learning is an effective way to improve LLM performance \cite{brown2020language} without updating weights at inference. Furthermore, \citet{wei2022chain} proposed the CoT prompting strategy in a few-shot setting, which we follow in our evaluation.

Additionally, we employ an automated annotation process, which enables us to annotate 100 samples from the training set to construct an exemplar pool. Since the annotation process does not contain the CoT process, we adopt GPT-4o to generate the reasoning steps. 

Given an atomic fact \(a_i\), we retrieve the most relevant few-shot exemplars \(\mathcal{E}\) by semantic matching between \(a_i\) and the exemplar pool. We then perform few-shot CoT fact verification conditioned on the dialogue history \(\mathcal{U}_{1:m-1}\), the external knowledge \(\mathcal{T}\), and the retrieved exemplars \(\mathcal{E}\), producing both a rationale and a veracity label:
%\textcolor{red}{We retrieve the most relevant samples $\mathcal{E}$ by semantically matching the atomic fact \(a_i\). The few-shot CoT fact verification given the dialogue history \(\mathcal{U}_{1:m-1}\), atomic fact \(a_i\), retrieved samples $\mathcal{E}$ and external knowledge \(\mathcal{T}\) is defined as follows}:

\begin{equation}
(o_i^{\text{fs},\text{reason}}, o_i^{\text{fs}, \text{label}}) = \mathcal{M}_{\theta} (p_{\text{fact}}, a_i, \mathcal{U}_{1:m-1}, \mathcal{T}, \mathcal{E}),
\end{equation}
where $o_i^{\text{fs},\text{reason}}$, $o_i^{\text{fs}, \text{label}}$ are few-shot LLM outputs. We use the same prompt template \(p_{\text{fact}}\) described in Section~\ref{Sec:Zero-Shot Chain of Thought} but add the exemplars for few-shot prompting.

\setlength{\tabcolsep}{2pt}
\renewcommand{\arraystretch}{0.95}
\begin{table*}[ht]
\centering
\resizebox{0.9\textwidth}{!}{%
\footnotesize
\begin{tabular}{l|cccccc|cccccc}
\toprule
\multirow{2}{*}{\textbf{Model}} & \multicolumn{6}{c|}{\textbf{HybriDialogue}} & \multicolumn{6}{c}{\textbf{OpenDialKG}} \\
\cmidrule(lr){2-7} \cmidrule(lr){8-13}
 & \textbf{Acc.} & \textbf{Prec.} & \textbf{Rec.} & \textbf{F1} & \textbf{Kappa} & \textbf{G-mean} & \textbf{Acc.} & \textbf{Prec.} & \textbf{Rec.} & \textbf{F1} & \textbf{Kappa} & \textbf{G-mean} \\
\midrule
\multicolumn{13}{c}{\textbf{Vanilla}} \\
\midrule
Mistral-7B-Instruct-v0.3 & 0.646 & 0.641 & 0.573 & 0.572 & 0.379 & 0.513 & 0.772 & 0.723 & 0.660 & 0.680 & 0.543 & 0.626 \\
Llama-3.1-8B-Instruct     & 0.557 & 0.493 & 0.496 & 0.429 & 0.208 & 0.242 & 0.714 & 0.672 & 0.594 & 0.542 & 0.377 & 0.300 \\
Llama-3.3-70B-Instruct    & 0.741 & 0.725 & \textbf{0.684} & 0.698 & 0.568 & \textbf{0.669} & 0.822 & 0.808 & 0.776 & 0.784 & 0.675 & 0.771 \\
Gemma-3-27B-it            & \textbf{0.757} & \textbf{0.750} & 0.682 & \textbf{0.699} & \textbf{0.591} & 0.652 & 0.831 & 0.799 & 0.774 & 0.780 & 0.691 & 0.768 \\
Qwen3-32B                 & 0.730 & 0.697 & 0.660 & 0.671 & 0.541 & 0.637 & 0.849 & \textbf{0.856} & 0.774 & 0.806 & 0.701 & 0.761 \\
QwQ-32B                   & 0.714 & 0.666 & 0.666 & 0.659 & 0.527 & 0.652 & \textbf{0.852} & 0.812 & \textbf{0.802} & \textbf{0.806} & \textbf{0.722} & \textbf{0.796} \\
GPT-4o                    & 0.711 & 0.666 & 0.665 & 0.665 & 0.524 & 0.656 & 0.788 & 0.745 & 0.783 & 0.760 & 0.627 & 0.782 \\
\midrule
\multicolumn{13}{c}{\textbf{CoT}} \\
\midrule
Mistral-7B-Instruct-v0.3          & 0.659 & 0.648 & 0.576 & 0.585 & 0.404 & 0.529 & 0.769 & 0.721 & 0.650 & 0.675 & 0.537 & 0.619 \\
DeepSeek-R1                           & 0.727 & 0.727 & 0.685 & 0.694 & 0.552 & 0.672 & 0.778 & 0.766 & 0.748 & 0.735 & 0.618 & 0.735 \\
Llama-3.1-8B-Instruct             & 0.651 & 0.596 & 0.515 & 0.493 & 0.389 & 0.311 & 0.668 & 0.574 & 0.489 & 0.474 & 0.355 & 0.322 \\
Llama-3.1-8B-Instruct$^{\diamond}$& 0.708 & 0.722 & 0.570 & 0.540 & 0.495 & 0.334 & 0.760 & 0.725 & 0.606 & 0.625 & 0.520 & 0.526 \\
Llama-3.1-8B-Instruct$^{*}$       & 0.759 & \textbf{0.759} & 0.666 & 0.671 & 0.597 & 0.594 & 0.782 & 0.759 & 0.700 & 0.694 & 0.611 & 0.655 \\
Llama-3.3-70B-Instruct            & 0.743 & 0.735 & 0.683 & 0.699 & 0.571 & 0.665 & 0.840 & \textbf{0.816} & 0.795 & 0.798 & 0.710 & 0.789 \\
Gemma-3-27B-it                    & \textbf{0.770} & 0.748 & 0.700 & 0.716 & \textbf{0.612} & 0.678 & 0.843 & 0.799 & 0.776 & 0.785 & 0.708 & 0.768 \\
Qwen3-32B                         & 0.759 & 0.719 & 0.687 & 0.696 & 0.599 & 0.660 & 0.849 & 0.812 & 0.819 & 0.812 & 0.728 & 0.817 \\
QwQ-32B                           & 0.759 & 0.716 & \textbf{0.725} & \textbf{0.720} & 0.607 & \textbf{0.720} & \textbf{0.855} & 0.807 & \textbf{0.844} & \textbf{0.824} & \textbf{0.740} & \textbf{0.843} \\
GPT-4o                            & 0.746 & 0.739 & 0.699 & 0.708 & 0.582 & 0.684 & 0.797 & 0.793 & 0.800 & 0.781 & 0.651 & 0.797 \\
\midrule
\multicolumn{13}{c}{\textbf{Few-Shot CoT}} \\
\midrule
Mistral-7B-Instruct-v0.3 & 0.643 & 0.677 & 0.577 & 0.574 & 0.372 & 0.514 & 0.766 & 0.740 & 0.671 & 0.672 & 0.529 & 0.619 \\
DeepSeek-R1                  & 0.719 & 0.732 & 0.669 & 0.680 & 0.538 & 0.647 & 0.769 & 0.787 & 0.746 & 0.735 & 0.605 & 0.732 \\
Llama-3.1-8B-Instruct     & 0.708 & 0.703 & 0.617 & 0.633 & 0.501 & 0.568 & 0.791 & 0.736 & 0.687 & 0.700 & 0.606 & 0.657 \\
Llama-3.3-70B-Instruct    & \textbf{0.776} & \textbf{0.752} & \textbf{0.730} & \textbf{0.740} & \textbf{0.627} & \textbf{0.722} & \textbf{0.862} & \textbf{0.833} & \textbf{0.828} & \textbf{0.830} & 0.744 & 0.825 \\
Gemma-3-27B-it            & \textbf{0.776} & 0.747 & 0.717 & 0.729 & 0.623 & 0.703 & 0.858 & 0.828 & 0.783 & 0.803 & 0.727 & 0.774 \\
Qwen3-32B                 & 0.724 & 0.711 & 0.663 & 0.672 & 0.544 & 0.636 & 0.825 & 0.799 & 0.805 & 0.796 & 0.688 & 0.804 \\
QwQ-32B                   & 0.768 & 0.721 & 0.716 & 0.718 & 0.616 & 0.705 & 0.858 & 0.793 & 0.825 & 0.807 & \textbf{0.745} & 0.824 \\
GPT-4o                    & 0.754 & 0.737 & 0.704 & 0.712 & 0.594 & 0.688 & 0.840 & 0.824 & 0.827 & 0.817 & 0.717 & \textbf{0.826} \\
\bottomrule
\end{tabular}}
\caption{Comparison of model performances on HybriDialogue and OpenDialKG datasets under Vanilla, CoT, and Few-Shot CoT settings. Kappa means Cohen’s kappa, indicating the inter-agreement between humans and models. Vanilla models refer to those without CoT reasoning. Llama-3.1-8B-Instruct$^{\diamond}$ is fine-tuned exclusively on factual labels, whereas Llama-3.1-8B-Instruct$^{*}$ is fine-tuned on factual labels augmented with reasoning steps. The best results are bolded in each category.}
\label{table:experimental_results}
\end{table*}

\subsection{Reasoning Distillation}
\label{Sec:Reasoning Distillation}
Traditional knowledge distillation processes knowledge, usually in the form of labels, from larger to smaller models. As we mentioned above, dialogue fact verification is more complex, and relying on teaching labels to smaller models is insufficient. 

Unlike the traditional method, we inject reasoning steps when distilling knowledge into student models. Specifically, we collect training samples by requesting GPT-4o to simulate the automated annotation process: identity verifiable factual claims, select the knowledge source, and generate the factual label with the reasoning steps.

After collecting these samples, We fine-tune the smaller models with LoRA \cite{hu2021lora}, which adapts the original weights via low-rank matrices. %LoRA is an efficient fine-tuning technique with a few extra parameters and lower computational resources. Another benefit is that it does not change the original LLM weight, fully leveraging the LLMs' strength. 
We optimise the student model using a combination of label-level and reasoning-level losses, defined as:
\begin{equation}
\mathcal{L} = \mathcal{L}_{\text{label}} + \mathcal{L}_{\text{reason}}.
\end{equation}

The label loss is the standard cross-entropy between the teacher-generated labels $y^{teacher}$ and the student predictions $p^{student}$:
\begin{equation}
\mathcal{L}_{\text{label}} = -\sum_{i=1}^{N} y_{i}^{\text{teacher}} \log p_{i}^{\text{student}},
\end{equation}
and the reasoning loss distils the reasoning sequences from the teacher to the student. We implement it as a token-level cross-entropy over the reasoning text generated by the teacher:
\begin{equation}
\mathcal{L}_{\text{reason}} = -\sum_{i=1}^{N} \sum_{j=1}^{L_i} \log P_{\theta}^{\text{student}}(w_{i,j} \mid w_{i,<j}),
\end{equation}
where $N$ is the number of samples, $L_i$ is the sequence length, $y$ is the ground-truth label, and $p$ is the predicted probability for the $i$-th sample. $w_{i,j}$ denotes the $j$-th token in the $i$-th reasoning sequence.

\setlength{\tabcolsep}{2pt}
\renewcommand{\arraystretch}{0.95}
\begin{table*}[ht]
\centering
\resizebox{0.9\textwidth}{!}{
\footnotesize
\begin{tabular}{l|cccccc|cccccc}
\toprule
\multirow{2}{*}{\textbf{Model}} & \multicolumn{6}{c|}{\textbf{HybriDialogue}} & \multicolumn{6}{c}{\textbf{OpenDialKG}} \\
\cmidrule(lr){2-7} \cmidrule(lr){8-13}
& \textbf{Acc.} & \textbf{Prec.} & \textbf{Rec.} & \textbf{F1} & \textbf{Kappa} & \textbf{G-mean} & \textbf{Acc.} & \textbf{Prec.} & \textbf{Rec.} & \textbf{F1} & \textbf{Kappa} & \textbf{G-mean} \\
\midrule
\multicolumn{13}{c}{\textbf{Vanilla}} \\
\midrule
Mistral-7B-Instruct-v0.3 & 0.631 & 0.666 & 0.654 & 0.585 & 0.355 & 0.581 & 0.677 & 0.660 & 0.669 & 0.580 & 0.393 & 0.597 \\
Llama-3.1-8B-Instruct    & 0.532 & 0.631 & 0.614 & 0.451 & 0.200 & 0.342 & 0.577 & 0.633 & 0.598 & 0.420 & 0.205 & 0.303 \\
Llama-3.3-70B-Instruct   & \textbf{0.855} & 0.816 & \textbf{0.851} & \textbf{0.827} & \textbf{0.743} & \textbf{0.847} & \textbf{0.811} & \textbf{0.741} & \textbf{0.804} & \textbf{0.744} & \textbf{0.650} & \textbf{0.791} \\
Gemma-3-27B-it           & 0.845 & \textbf{0.818} & 0.846 & 0.822 & 0.726 & 0.840 & \textbf{0.811} & 0.737 & 0.789 & 0.737 & 0.649 & 0.776 \\
QwQ-32B                  & 0.751 & 0.700 & 0.786 & 0.684 & 0.578 & 0.758 & 0.710 & 0.665 & 0.718 & 0.600 & 0.474 & 0.660 \\
Qwen3-32B                & 0.758 & 0.763 & 0.777 & 0.737 & 0.573 & 0.752 & 0.753 & 0.733 & 0.736 & 0.689 & 0.532 & 0.703 \\
\midrule
\multicolumn{13}{c}{\textbf{CoT}} \\
\midrule
Mistral-7B-Instruct-v0.3          & 0.654 & 0.663 & 0.649 & 0.602 & 0.390 & 0.597 & 0.698 & 0.667 & 0.679 & 0.608 & 0.431 & 0.625 \\
Deepseek-R1                          & 0.871 & 0.828 & 0.858 & 0.842 & 0.769 & 0.858 & \textbf{0.856} & \textbf{0.782} & 0.828 & \textbf{0.801} & 0.736 & 0.826 \\
Llama-3.1-8B-Instruct             & 0.740 & 0.690 & 0.579 & 0.594 & 0.519 & 0.471 & 0.734 & 0.637 & 0.543 & 0.556 & 0.479 & 0.419 \\
Llama-3.1-8B-Instruct$^{\diamond}$& 0.727 & 0.793 & 0.545 & 0.553 & 0.490 & 0.378 & 0.766 & 0.735 & 0.641 & 0.666 & 0.548 & 0.600 \\
Llama-3.1-8B-Instruct$^{*}$       & 0.831 & 0.820 & 0.776 & 0.795 & 0.694 & 0.769 & 0.817 & 0.765 & 0.755 & 0.759 & 0.658 & 0.748 \\
Llama-3.3-70B-Instruct            & \textbf{0.872} & 0.828 & \textbf{0.874} & \textbf{0.843} & \textbf{0.773} & \textbf{0.871} & 0.839 & 0.776 & 0.832 & 0.785 & \textbf{0.702} & 0.824 \\
Gemma-3-27B-it                    & 0.854 & \textbf{0.831} & 0.870 & 0.836 & 0.741 & 0.862 & 0.824 & 0.769 & 0.802 & 0.764 & 0.671 & 0.789 \\
QwQ-32B                           & 0.843 & 0.768 & 0.872 & 0.789 & 0.729 & 0.865 & 0.828 & 0.750 & \textbf{0.846} & 0.760 & 0.687 & \textbf{0.833} \\
Qwen3-32B                         & 0.842 & 0.793 & 0.847 & 0.811 & 0.721 & 0.843 & 0.816 & 0.757 & 0.815 & 0.770 & 0.660 & 0.807 \\
\midrule
\multicolumn{13}{c}{\textbf{Few-Shot CoT}} \\
\midrule
Mistral-7B-Instruct-v0.3 & 0.590 & 0.625 & 0.640 & 0.531 & 0.296 & 0.537 & 0.646 & 0.637 & 0.634 & 0.539 & 0.332 & 0.542 \\
Deepseek-R1                 & 0.873 & 0.832 & 0.859 & \textbf{0.845} & \textbf{0.774} & 0.859 & \textbf{0.857} & \textbf{0.802} & 0.817 & \textbf{0.809} & \textbf{0.736} & 0.814 \\
Llama-3.1-8B-Instruct    & 0.816 & 0.806 & 0.790 & 0.788 & 0.671 & 0.781 & 0.801 & 0.740 & 0.750 & 0.730 & 0.626 & 0.736 \\
Llama-3.3-70B-Instruct   & 0.853 & \textbf{0.812} & 0.862 & 0.824 & 0.741 & 0.856 & 0.831 & 0.776 & 0.827 & 0.782 & 0.685 & 0.816 \\
Gemma-3-27B-it           & 0.828 & 0.804 & 0.843 & 0.802 & 0.697 & 0.831 & 0.805 & 0.774 & 0.797 & 0.757 & 0.633 & 0.778 \\
QwQ-32B                  & \textbf{0.856} & 0.795 & \textbf{0.878} & 0.817 & 0.749 & \textbf{0.872} & 0.828 & 0.759 & \textbf{0.833} & 0.766 & 0.683 & \textbf{0.822} \\
Qwen3-32B                & 0.853 & 0.805 & 0.843 & 0.821 & 0.738 & 0.842 & 0.827 & 0.767 & 0.817 & 0.782 & 0.680 & 0.812 \\
\bottomrule
\end{tabular}}
\caption{Comparison of model performance on the GPT-4o–annotated HybriDialogue and OpenDialKG datasets. Llama-3.1-8B-Instruct$^{\diamond}$ is fine-tuned solely on factual labels, while Llama-3.1-8B-Instruct$^{*}$ is fine-tuned on factual labels enriched with reasoning steps. We bold the best result in each category.}
\label{table:experimental_results_GPT_annotated}
\end{table*}

\begin{table*}[ht]
\centering
\resizebox{0.9\textwidth}{!}{%
\setlength{\tabcolsep}{3pt}
\footnotesize
\begin{tabular}{l|l|ccc|ccc}
\toprule
\multirow{2}{*}{\textbf{Mode}} & \multirow{2}{*}{\textbf{Model}} 
& \multicolumn{3}{c|}{\textbf{Response-level}} 
& \multicolumn{3}{c}{\textbf{Atomic-Fact-level}} \\
\cmidrule(lr){3-5}\cmidrule(lr){6-8}
 &  & \textbf{SUPPORTS} & \textbf{REFUTES} & \textbf{NEI} & \textbf{SUPPORTS} & \textbf{REFUTES} & \textbf{NEI} \\
\midrule
\multicolumn{8}{c}{\textbf{OpenDialKG Dataset}} \\
\midrule
Vanilla        & Mistral-7B-Instruct-v0.3  & 441 (0.641) & 142 (0.206) & 105 (0.153) & 3137 (0.769) & 394 (0.097) & 546 (0.134) \\
Few-Shot CoT & Qwen3-32B                 & 271 (0.394) & 145 (0.211) & 272 (0.395) & 2423 (0.594) & 277 (0.068) & 1377 (0.338) \\
Few-Shot CoT & Llama-3.3-70B-Instruct    & 288 (0.419) & 159 (0.231) & 241 (0.350) & 2578 (0.632) & 297 (0.073) & 1202 (0.295) \\
\midrule
\multicolumn{8}{c}{\textbf{HybriDialogue Dataset}} \\
\midrule
Vanilla        & Mistral-7B-Instruct-v0.3  & 658 (0.629) & 173 (0.165) & 215 (0.206) & 3027 (0.761) & 350 (0.088) & 601 (0.151) \\
Few-Shot CoT & Qwen3-32B                 & 335 (0.320) & 166 (0.159) & 545 (0.521) & 1944 (0.489) & 299 (0.075) & 1735 (0.436) \\
Few-Shot CoT & Llama-3.3-70B-Instruct    & 399 (0.381) & 183 (0.175) & 464 (0.444) & 2205 (0.554) & 330 (0.083) & 1443 (0.363) \\
\bottomrule
\end{tabular}}
\caption{Number and proportions of \emph{Supports}, \emph{Refutes}, and \emph{NEI (Not Enough Information)} predicted at two levels by several LLMs on the GPT-4o-annotated HybriDialogue and OpenDialKG datasets.}
\label{table:response_atomic_comparison_updated}
\end{table*}

\section{Experiment}

\subsection{Baselines}
We adopt several LLMs as baselines with various baseline methods to measure the performance of models. The LLMs include Mistral-7B \cite{albert2023mistral}, Llama3 \cite{dubey2024llama}, Deepseek-R1 \cite{guo2025deepseek}, Qwen3 \cite{yang2025qwen3}, QwQ \cite{team2025qwq}, Gemma3 \cite{team2025gemma}, Gemini \cite{team2024gemini}, GPT \cite{hurst2024gpt}.

\subsection{Experimental Setup}
The samples used for fine-tuning are from the train set of OpenDialKG and generated by GPT-4o, with a total of 1429. We used LoRA to fine-tune our smaller language models, with the settings of rank 32 and alpha 32. We fine-tune the Llama 3 8B models for 3 epochs with a single 80GB A100 GPU. For the open-source models, the inference with the Llama 3 70B models requires two 80GB A100 GPUs, and all the other models use one. The experiments were conducted using a fixed random seed of 42, with a single run.

\begin{table*}[ht]
\centering
\resizebox{0.95\textwidth}{!}{%
\begin{tabular}{@{}p{4.5cm}p{15cm}@{}}
\toprule
\textbf{Case Study} & \\
\midrule
\textbf{Evidence} & John Corson (born April 4, 1971), better known by the ring name John Zandig, is an American professional wrestler and promoter. He is the founder and former owner of the promotion Combat Zone Wrestling (CZW). \\

\midrule

\textbf{Speaker A} & Do you know anything about CZW's Cage of Death? \\

\textbf{Speaker B} & Yes, it is a event that has been around since 1999. A steel cage with weapons is typically found and winners win through pin or use of weapons. There has been different format types as well such as WarGames in Cage of Death 5, 6 and 7. \\
%\textbf{Dialogue History} & ...\\

\textbf{Speaker A} & Oh wow that sounds pretty brutal. There seems to be a lot but I heard stuff about the one in 2002, was there anyone notable there? \\

\textbf{Speaker B} & Yes, Cage of Death 4 was held in 2002. Notable people would be John Zandig defeating Lobo for the owner of CZW. \\

\textbf{Atomic Fact (one of)} & John Zandig is the owner of CZW. \\

\midrule

\textbf{Qwen3-32B CoT} & \textbf{Decision : Refutes } 

\emph{chain of thought}:
let's evaluate the statement ``john zandig is the owner of czw'' based on the provided evidence.
1. the evidence states: ``john corson (born april 4, 1971), better known by the ring name john zandig, is an american professional wrestler and promoter. he is the founder and former owner of the promotion combat zone wrestling (czw).''
2. the statement in question is ``john zandig is the owner of czw.''
3. the evidence says he is the ``founder and former owner'' of czw. this implies that at one time, he was the owner, but it does not confirm that he is currently the owner.
4. the statement claims he is the owner, but the evidence only confirms he was the owner in the past.
conclusion: the evidence does not directly support the current claim that john zandig is the owner of czw. it only states he was the former owner.
final answer: false \\
\midrule

\textbf{Qwen3-32B Few-Shot CoT} & \textbf{Decision: Supports}

\emph{chain of thought}:
1. the statement claims that john zandig is the owner of czw.
2. there are no pronouns or references to resolve in the statement.
3. the knowledge source states that john zandig is the founder and former owner of the promotion combat zone wrestling (czw).
4. the term ``former owner'' indicates that he is no longer the current owner, but the statement does not specify a time frame and only asserts that he is the owner.
5. since the knowledge source directly mentions that he was the owner, and the statement does not contradict this past fact, the statement is supported by the evidence.
output: true \\
\midrule
\textbf{Ground truth} & \textbf{Refutes}\\

\bottomrule
\end{tabular}}
\caption{Case study comparing Qwen3-32B under CoT and few-shot CoT prompting strategies.}
\label{table:case_study_qwen3}
\end{table*}

\subsection{Results on Human-Annotated Datasets}
We report dialogue fact verification results on two manually annotated datasets, OpenDialKG and HybriDialogue (Table~\ref{table:experimental_results}).

In the vanilla setting, Llama-3.3-70B-Instruct and Gemma-3-27B-it deliver the best overall results across both datasets. On HybriDialogue, Gemma-3-27B achieves the highest accuracy and Cohen’s kappa, while on OpenDialKG, QwQ-32B achieves the best F1 score, indicating stronger generalisation on that dataset.

With CoT prompting, all models improve across metrics on both datasets, particularly recall and Cohen’s kappa. Under CoT, Gemma-3-27B-it and QwQ-32B are the top performers; QwQ-32B yields the highest F1 (0.720 on HybriDialogue and 0.740 on OpenDialKG), demonstrating that CoT benefits LLM understanding. Generally, a higher G-mean with CoT indicates a more balanced per-class recall under class imbalance, suggesting improvements are not driven solely by the majority class. 

Regarding distillation, Llama-3.1-8B-Instruct$^{\diamond}$ denotes that it is distilled by factual label only, and Llama-3.1-8B-Instruct$^{*}$ denotes that it is distilled by factual label and reasoning step. Llama-3.1-8B-Instruct models with distillation both outperform Llama-3.1-8B-Instruct, but we observe a larger improvement with the reasoning step. It indicates the efficiency of reasoning distillation.

Under few-shot CoT, the Llama series, Gemma-3-27B-it, and GPT-4o further improve over the CoT setting. Llama-3.3-70B-Instruct, in particular, improves markedly, reaching F1 scores of 0.740 and 0.830. These gains suggest that few-shot examples are especially helpful for models without explicit reasoning capabilities. However, we also see a drop in some models, like Qwen3-32B, QwQ-32B. After analysis, it is caused by over-reasoning, which we will discuss more in Section~\ref{sec:case_study}.

\subsection{Results on GPT-4o-Annotated Datasets}
Table~\ref{table:experimental_results_GPT_annotated} presents results on the automated datasets.

Overall trends mirror those observed on the human-annotated datasets: performance increases after applying CoT and few-shot CoT. Two differences stand out. First, scores on HybriDialogue are generally higher than their human-annotated counterparts, indicating stronger alignment between the evaluated LLMs and GPT-4o annotations. Second, the improvement pattern diverges: under few-shot settings, the performance of the Llama series and Gemma-3-27B-it degrades compared to CoT, in contrast to the gains observed on human-annotated datasets. After analysis, the main reason is also caused by over-reasoning.

\subsection{Granularity Effects on Predicted Label Distribution}
We analyse the effects on the predicted label distribution at both the response level and the atomic-fact level, where the data annotation for the response level is discussed in Section~\ref{sec:automated_annotation}.

\begin{figure}[t]
\centering
  \includegraphics[width=1\linewidth]{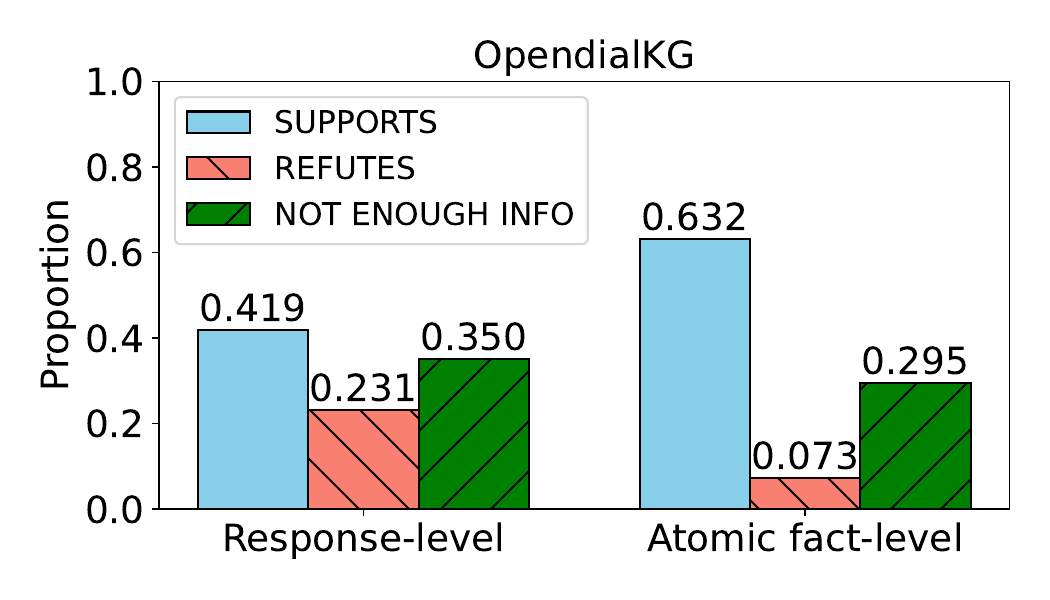}
  \caption {Proportions of \emph{Supports}, \emph{Refutes}, and \emph{Not Enough Information} predicted at two levels by Llama-3.3-70B-Instruct on the OpenDialKG dataset.}
  \label{figure:distribution_response}
\end{figure}

We adopt the same setting to verify dialogue facts at the response level and report the predicted label distribution, as shown in Table~\ref{table:response_atomic_comparison_updated}. The distribution shows a similar trend: the proportion of responses at the \emph{Supports} level is lower than at the atomic-fact level, while the proportion of \emph{Refutes} and \emph{Not Enough Information} is higher (Figure~\ref{figure:distribution_response} illustrates this trend). These can be attributed to reasons: coarse-grained evidence retrieval and judgment.

In summary, focusing on the dialogue response level is not reliable, as it predicts higher \emph{Refutes} and cannot reflect the actual hallucinations in dialogue. 

\subsection{Case Study}
\label{sec:case_study}
We present a case study on Qwen3 using both CoT and few-shot CoT prompting, as shown in Table~\ref{table:case_study_qwen3}. Under the CoT setting, Qwen3 demonstrates strong reasoning ability and successfully arrives at the correct conclusion. In contrast, when few-shot examples are provided, Qwen3 tends to relax the temporal constraint and make an unsupported inference that ``is the owner'' can be satisfied by ``was the owner,'' thus incorrectly predicting \emph{Supports}. The case suggests that few-shot examples may induce over-justification, which does not consistently improve performance.

\subsection{Discussion}
Our experimental results show that CoT, few-shot CoT and CoT reasoning benefit improving LLMs' performance in dialogue fact verification. We summarise the common errors as coarse-grained evidence selection and judgment at the response level, and over-justification in the CoT reasoning. Coarse-grained evidence arises from focusing primarily on the response level, which makes it difficult to comprehensively identify supporting evidence, as responses often contain a mix of factual and non-factual information. Coarse-grained judgment is influenced by the diversity of factual content within a response; models tend to classify responses as refuted even when they contain accurate information. We observe over-justification in few-shot CoT prompting, where the model over-interprets specific details and reaches conclusions that conflict with human annotations.

\section{Conclusion}
We propose FineDialFact, a novel benchmark dataset for fine-grained fact verification in dialogue to address the limitations that
previous fact verification on dialogue focused on the response level, which is coarse-grained. To verify dialogue facts in a fine-grained way, we process the response into verifiable atomic facts, enabling the challenging yet realistic scenario where different facts within a dialogue can have different factual labels.
Given that no existing related datasets were available, we constructed the dataset by collecting hallucinated samples, splitting responses into atomic facts, retrieving knowledge, recruiting participants for manual annotation, and enlarging the dataset through automated annotation.

We also perform benchmarking experiments with CoT baselines. Experimental results show that CoT can significantly improve the models' performance, and reasoning distillation is a useful method for helping smaller models achieve strong performance. It also shows that the task is far from solved. On the human-annotated HybriDialogue dataset, the highest F1-score achieved is 0.74, indicating that dialogue fact verification is still challenging. 

By labelling claims as verifiable or non-verifiable, our dataset enables exploration of the relationship between verifiable claim detection and verification, paving the way toward a more unified and comprehensive fact-checking pipeline.

%In addition, our current knowledge base relies exclusively on Wikipedia, which presents a limitation, as incorporating additional sources could enhance the robustness of the verification process and could be our future work to explore.

\section*{Limitations}
While our proposed benchmark makes a significant contribution to fine-grained dialogue fact verification enabling research in a task lacking datasets to date, our work has some limitations.

Our current knowledge base relies exclusively on Wikipedia, which may present limitations where the evidence is not available and additional knowledge bases may be needed. Hence, incorporating additional sources could enhance the robustness of the verification process.

Despite providing a first-of-its-kind resource, FineDialFact is limited in size to 1,000 human-annotated samples. Our research not only provides a resource, but also a methodology to develop similar resources, and therefore we hope to motivate the research community and see more datasets of this kind in the near future.

We split the dialogue response into several pieces of atomic facts to verify, gaining more accurate results. As a caveat, rather than as a limitation per se, this increases the cost of using GPUs, increasing the computational resources needed to perform this research, which we argue however that it is necessary and beneficial.

\section*{Ethical Statement}
Our work involves human annotations; however, the tasks were limited to labelling a predefined range of options, such as selecting factual labels, and did not involve the collection or use of any personal information.

The datasets we used, HybriDialogue and OpendialKG, are publicly available, and no additional personally sensitive information was added in our benchmark.

\section*{Acknowledgments}
We acknowledge financial support from UKRI through the grant Responsible AI UK (EP/Y009800/1) keystone project AdSoLve (KP0016); from the Slovenian Research Agency ARIS via the project  LLM4DH (GC-0002) and research core funding for the programme Knowledge Technologies (P2-0103); and from the European Union’s Horizon Europe research and innovation programme under grant agreement No 101214398 (ELLIOT). Views and opinions expressed are however those of the author(s) only and do not necessarily reflect those of the European Union or the European Commission. Neither the European Union nor the European Commission can be held responsible for them.

\section{Bibliographical References}\label{sec:reference}

\bibliographystyle{lrec2026-natbib}
\bibliography{lrec2026-example}

@article{chung2024scaling,
  title={Scaling instruction-finetuned language models},
  author={Chung, Hyung Won and Hou, Le and Longpre, Shayne and Zoph, Barret and Tay, Yi and Fedus, William and Li, Yunxuan and Wang, Xuezhi and Dehghani, Mostafa and Brahma, Siddhartha and others},
  journal={Journal of Machine Learning Research},
  volume={25},
  number={70},
  pages={1--53},
  year={2024}
}

@article{chen2024diahalu,
  title={DiaHalu: A Dialogue-level Hallucination Evaluation Benchmark for Large Language Models},
  author={Chen, Kedi and Chen, Qin and Zhou, Jie and He, Yishen and He, Liang},
  journal={arXiv preprint arXiv:2403.00896},
  year={2024}
}

@article{min2023factscore,
  title={Factscore: Fine-grained atomic evaluation of factual precision in long form text generation},
  author={Min, Sewon and Krishna, Kalpesh and Lyu, Xinxi and Lewis, Mike and Yih, Wen-tau and Koh, Pang Wei and Iyyer, Mohit and Zettlemoyer, Luke and Hajishirzi, Hannaneh},
  journal={arXiv preprint arXiv:2305.14251},
  year={2023}
}

@article{song2024finesure,
  title={FineSurE: Fine-grained summarization evaluation using LLMs},
  author={Song, Hwanjun and Su, Hang and Shalyminov, Igor and Cai, Jason and Mansour, Saab},
  journal={arXiv preprint arXiv:2407.00908},
  year={2024}
}

@inproceedings{yu2022xdai,
  title={XDAI: A tuning-free framework for exploiting pre-trained language models in knowledge grounded dialogue generation},
  author={Yu, Jifan and Zhang, Xiaohan and Xu, Yifan and Lei, Xuanyu and Guan, Xinyu and Zhang, Jing and Hou, Lei and Li, Juanzi and Tang, Jie},
  booktitle={Proceedings of the 28th ACM SIGKDD Conference on Knowledge Discovery and Data Mining},
  pages={4422--4432},
  year={2022}
}

@article{izacard2021unsupervised,
  title={Unsupervised dense information retrieval with contrastive learning},
  author={Izacard, Gautier and Caron, Mathilde and Hosseini, Lucas and Riedel, Sebastian and Bojanowski, Piotr and Joulin, Armand and Grave, Edouard},
  journal={arXiv preprint arXiv:2112.09118},
  year={2021}
}

@article{gupta2021dialfact,
  title={DialFact: A benchmark for fact-checking in dialogue},
  author={Gupta, Prakhar and Wu, Chien-Sheng and Liu, Wenhao and Xiong, Caiming},
  journal={arXiv preprint arXiv:2110.08222},
  year={2021}
}

@article{dubey2024llama,
  title={The llama 3 herd of models},
  author={Dubey, Abhimanyu and Jauhri, Abhinav and Pandey, Abhinav and Kadian, Abhishek and Al-Dahle, Ahmad and Letman, Aiesha and Mathur, Akhil and Schelten, Alan and Yang, Amy and Fan, Angela and others},
  journal={arXiv preprint arXiv:2407.21783},
  year={2024}
}

@inproceedings{ni2023multi,
  title={Multi-source multi-type knowledge exploration and exploitation for dialogue generation},
  author={Ni, Xuanfan and Dai, Hongliang and Ren, Zhaochun and Li, Piji},
  booktitle={Proceedings of the 2023 Conference on Empirical Methods in Natural Language Processing},
  pages={12522--12537},
  year={2023}
}

@article{li2022eliciting,
  title={Eliciting knowledge from large pre-trained models for unsupervised knowledge-grounded conversation},
  author={Li, Yanyang and Zhao, Jianqiao and Lyu, Michael R and Wang, Liwei},
  journal={arXiv preprint arXiv:2211.01587},
  year={2022}
}

@article{shuster2021retrieval,
  title={Retrieval augmentation reduces hallucination in conversation},
  author={Shuster, Kurt and Poff, Spencer and Chen, Moya and Kiela, Douwe and Weston, Jason},
  journal={arXiv preprint arXiv:2104.07567},
  year={2021}
}

@article{brown2020language,
  title={Language models are few-shot learners},
  author={Brown, Tom and Mann, Benjamin and Ryder, Nick and Subbiah, Melanie and Kaplan, Jared D and Dhariwal, Prafulla and Neelakantan, Arvind and Shyam, Pranav and Sastry, Girish and Askell, Amanda and others},
  journal={Advances in neural information processing systems},
  volume={33},
  pages={1877--1901},
  year={2020}
}

@article{wei2022chain,
  title={Chain-of-thought prompting elicits reasoning in large language models},
  author={Wei, Jason and Wang, Xuezhi and Schuurmans, Dale and Bosma, Maarten and Xia, Fei and Chi, Ed and Le, Quoc V and Zhou, Denny and others},
  journal={Advances in neural information processing systems},
  volume={35},
  pages={24824--24837},
  year={2022}
}

@inproceedings{moon2019opendialkg,
  title={Opendialkg: Explainable conversational reasoning with attention-based walks over knowledge graphs},
  author={Moon, Seungwhan and Shah, Pararth and Kumar, Anuj and Subba, Rajen},
  booktitle={Proceedings of the 57th annual meeting of the association for computational linguistics},
  pages={845--854},
  year={2019}
}

@article{nakamura2022hybridialogue,
  title={HybriDialogue: An information-seeking dialogue dataset grounded on tabular and textual data},
  author={Nakamura, Kai and Levy, Sharon and Tuan, Yi-Lin and Chen, Wenhu and Wang, William Yang},
  journal={arXiv preprint arXiv:2204.13243},
  year={2022}
}

@article{hurst2024gpt,
  title={Gpt-4o system card},
  author={Hurst, Aaron and Lerer, Adam and Goucher, Adam P and Perelman, Adam and Ramesh, Aditya and Clark, Aidan and Ostrow, AJ and Welihinda, Akila and Hayes, Alan and Radford, Alec and others},
  journal={arXiv preprint arXiv:2410.21276},
  year={2024}
}

@article{zhang2022automatic,
  title={Automatic chain of thought prompting in large language models},
  author={Zhang, Zhuosheng and Zhang, Aston and Li, Mu and Smola, Alex},
  journal={arXiv preprint arXiv:2210.03493},
  year={2022}
}

@article{li2023symbolic,
  title={Symbolic chain-of-thought distillation: Small models can also" think" step-by-step},
  author={Li, Liunian Harold and Hessel, Jack and Yu, Youngjae and Ren, Xiang and Chang, Kai-Wei and Choi, Yejin},
  journal={arXiv preprint arXiv:2306.14050},
  year={2023}
}

@article{kojima2022large,
  title={Large language models are zero-shot reasoners},
  author={Kojima, Takeshi and Gu, Shixiang Shane and Reid, Machel and Matsuo, Yutaka and Iwasawa, Yusuke},
  journal={Advances in neural information processing systems},
  volume={35},
  pages={22199--22213},
  year={2022}
}

@article{cohen1960coefficient,
  title={A coefficient of agreement for nominal scales},
  author={Cohen, Jacob},
  journal={Educational and psychological measurement},
  volume={20},
  number={1},
  pages={37--46},
  year={1960},
  publisher={Sage Publications Sage CA: Thousand Oaks, CA}
}

@article{hu2021lora,
  title={Lora: Low-rank adaptation of large language models},
  author={Hu, Edward J and Shen, Yelong and Wallis, Phillip and Allen-Zhu, Zeyuan and Li, Yuanzhi and Wang, Shean and Wang, Lu and Chen, Weizhu},
  journal={arXiv preprint arXiv:2106.09685},
  year={2021}
}

@article{jaccard1901etude,
  title={{\'E}tude comparative de la distribution florale dans une portion des Alpes et des Jura},
  author={Jaccard, Paul},
  journal={Bull Soc Vaudoise Sci Nat},
  volume={37},
  pages={547--579},
  year={1901}
}

@inproceedings{wan2024acueval,
  title={ACUEval: Fine-grained hallucination evaluation and correction for abstractive summarization},
  author={Wan, David and Sinha, Koustuv and Iyer, Srini and Celikyilmaz, Asli and Bansal, Mohit and Pasunuru, Ramakanth},
  booktitle={Findings of the Association for Computational Linguistics ACL 2024},
  pages={10036--10056},
  year={2024}
}

@inproceedings{tan2019recognizing,
  title={Recognizing conflict opinions in aspect-level sentiment classification with dual attention networks},
  author={Tan, Xingwei and Cai, Yi and Zhu, Changxi},
  booktitle={Proceedings of the 2019 conference on empirical methods in natural language processing and the 9th international joint conference on natural language processing (EMNLP-IJCNLP)},
  pages={3426--3431},
  year={2019}
}

@article{farquhar2024detecting,
  title={Detecting hallucinations in large language models using semantic entropy},
  author={Farquhar, Sebastian and Kossen, Jannik and Kuhn, Lorenz and Gal, Yarin},
  journal={Nature},
  volume={630},
  number={8017},
  pages={625--630},
  year={2024},
  publisher={Nature Publishing Group UK London}
}

@article{yang2025qwen3,
  title={Qwen3 technical report},
  author={Yang, An and Li, Anfeng and Yang, Baosong and Zhang, Beichen and Hui, Binyuan and Zheng, Bo and Yu, Bowen and Gao, Chang and Huang, Chengen and Lv, Chenxu and others},
  journal={arXiv preprint arXiv:2505.09388},
  year={2025}
}

@misc{team2025qwq,
  title={Qwq-32b: Embracing the power of reinforcement learning},
  author={Team, Qwen},
  year={2025},
  month=mar,
  howpublished = {\url{https://qwenlm.github.io/blog/qwq-32b}},
}

@article{team2025gemma,
  title={Gemma 3 technical report},
  author={Team, Gemma and Kamath, Aishwarya and Ferret, Johan and Pathak, Shreya and Vieillard, Nino and Merhej, Ramona and Perrin, Sarah and Matejovicova, Tatiana and Ram{\'e}, Alexandre and Rivi{\`e}re, Morgane and others},
  journal={arXiv preprint arXiv:2503.19786},
  year={2025}
}

@article{team2024gemini,
  title={Gemini 1.5: Unlocking multimodal understanding across millions of tokens of context},
  author={Team, Gemini and Georgiev, Petko and Lei, Ving Ian and Burnell, Ryan and Bai, Libin and Gulati, Anmol and Tanzer, Garrett and Vincent, Damien and Pan, Zhufeng and Wang, Shibo and others},
  journal={arXiv preprint arXiv:2403.05530},
  year={2024}
}

@article{guo2025deepseek,
  title={Deepseek-r1: Incentivizing reasoning capability in llms via reinforcement learning},
  author={Guo, Daya and Yang, Dejian and Zhang, Haowei and Song, Junxiao and Zhang, Ruoyu and Xu, Runxin and Zhu, Qihao and Ma, Shirong and Wang, Peiyi and Bi, Xiao and others},
  journal={arXiv preprint arXiv:2501.12948},
  year={2025}
}

@article{albert2023mistral,
  title={Mistral 7B},
  author={Albert, Q Jiang and Sablayrolles, Alexandre and Mensch, Arthur and Bamford, Chris and Chaplot, Devendra Singh},
  journal={arXiv preprint},
  year={2023}
}

@article{mitra2024factlens,
  title={FactLens: Benchmarking Fine-Grained Fact Verification},
  author={Mitra, Kushan and Zhang, Dan and Rahman, Sajjadur and Hruschka, Estevam},
  journal={arXiv preprint arXiv:2411.05980},
  year={2024}
}

@article{zhao2023survey,
  title={A survey of large language models},
  author={Zhao, Wayne Xin and Zhou, Kun and Li, Junyi and Tang, Tianyi and Wang, Xiaolei and Hou, Yupeng and Min, Yingqian and Zhang, Beichen and Zhang, Junjie and Dong, Zican and others},
  journal={arXiv preprint arXiv:2303.18223},
  volume={1},
  number={2},
  year={2023}
}

@article{ji2023survey,
  title={Survey of hallucination in natural language generation},
  author={Ji, Ziwei and Lee, Nayeon and Frieske, Rita and Yu, Tiezheng and Su, Dan and Xu, Yan and Ishii, Etsuko and Bang, Ye Jin and Madotto, Andrea and Fung, Pascale},
  journal={ACM computing surveys},
  volume={55},
  number={12},
  pages={1--38},
  year={2023},
  publisher={ACM New York, NY}
}

@book{robertson2009probabilistic,
  title={The probabilistic relevance framework: BM25 and beyond},
  author={Robertson, Stephen and Zaragoza, Hugo},
  volume={4},
  year={2009},
  publisher={Now Publishers Inc}
}

%\label{lr:ref}
%\bibliographystylelanguageresource{lrec2026-natbib}
%\bibliographylanguageresource{languageresource}
%\include{LREC2026 Author's kit/appendix}
\newpage

\appendix
\section{Prompts}
\label{Appendix:prompts}
The prompts for dialogue response generation, atomic facts splitting and dialogue fact verification are listed in these tables~\ref{table:prompt_response_generation}, ~\ref{table:prompt_atomic_fact_splitting}, and ~\ref{table:prompt_fact_verfication}.

\begin{table}[h]
\renewcommand{\arraystretch}{1.5}
\begin{tabular}{p{7.6cm}}
\toprule
\textbf{Prompt for Dialogue Response Generation} \\
\midrule
\midrule
\textbf{Dialogue:} \{Dialogue History\}

\textbf{Instruction:} 

Given the above dialogue, please respond to the input below and ensure the response is fluent and fact-consistent in English. 

\textbf{Input:} \{The utterance of Speaker A\}

\textbf{Response:} \\

\bottomrule
\end{tabular}
\caption{The prompt for dialogue response generation.}
\label{table:prompt_response_generation}
\end{table}

\begin{table}[h]
\renewcommand{\arraystretch}{1.5}
\begin{tabular}{p{7.6cm}}
\toprule
\textbf{Prompt for Atomic Fact Splitting} \\
\midrule
\midrule
\textbf{Examples:} \{Retrieved Examples\}

If the following input is an incomplete sentence or a phrase, please output it exactly as it is. Otherwise, if it is a complete sentence, split it into atomic sentences based only on the given information, without adding any additional information or making inferences.

\textbf{Input:} \{Response\}

\textbf{Output:} \{Atomic facts\} \\
\bottomrule
\end{tabular}
\caption{The prompt for atomic fact splitting.}
\label{table:prompt_atomic_fact_splitting}
\end{table}

\begin{table*}[h]
\renewcommand{\arraystretch}{1.5}
\begin{tabular}{p{15.5cm}}
\toprule
\textbf{Prompt for Dialogue Fact Verification} \\
\midrule
\midrule
\{Demonstrations\} \\
\textbf{Instruction:} \\
The statement is part of a response in a dialogue. Evaluate the statement strictly based on the provided knowledge source and dialogue history only. 

If the statement is not a factual claim (e.g., opinion, question, or unclear assertion), output: \texttt{``not enough information.''} 

If it is a factual claim:
\begin{itemize}
  \item Output \textbf{true} if the statement is directly supported by evidence in the knowledge source or dialogue history.
  \item Output \textbf{false} if the statement is directly contradicted by the knowledge source or dialogue history.
  \item Output \textbf{not enough information} if there is no direct evidence for or against the statement.
\end{itemize} \\

\textbf{Important:} 

Do not use your internal knowledge or make inferences.

Please think step by step and output your final answer. 

\textbf{Evidence:} \{Knowledge Source\} 

\textbf{Dialogue History:} \{Dialogue History\} 

\textbf{Statement:} \{Atomic Fact\} 

\textbf{Output:} \\
\bottomrule
\end{tabular}
\caption{The prompt for our dialogue fact verification. The prompt can be used for vanilla, CoT and few-shot CoT by adjusting the prompt slightly.}
\label{table:prompt_fact_verfication}
\end{table*}

\section{Annotation Instruction}
The details of the annotation instruction are listed in Table~\ref{table:factual_annotation_instructions}. Before annotation, we fully informed the participants that the annotated data would be used in our research and obtained their consent.

\label{Sec:annotation-instruction}
\begin{table*}[h]
\renewcommand{\arraystretch}{1.5}
\begin{tabular}{p{15.5cm}}
\toprule
\textbf{Human Annotation Instructions} \\
\midrule
\midrule
The task aims at annotating dialogue factual responses. For each sample, we provide you with a dialogue, several pieces of evidence, and two labels—factual claim and factual label. Your task is to select the most relevant pieces of evidence (as much as possible) and determine the labels.

There is a list of samples containing dialogue and evidence. Our goal is to select evidence for the last utterance and identify if the last utterance is verifiable or non-verifiable. You need to use the annotation tool to:

\textbf{1. Factual Claim Discrimination}

First, you have to determine whether the last utterance is a factual claim. A factual claim normally contains:
\begin{itemize}
  \item Specific, verifiable information that can be proven true or false
  \item Statements about events, measurements, statistics, or observable phenomena
  \item References to dates, times, people, places, or quantities
  \item Content that could be checked against reliable sources or evidence
  \item Statements that are objective rather than expressing opinions or preferences
\end{itemize}

If it is a factual claim, select \textbf{[Verifiable]} and proceed to step 2. Otherwise, select \textbf{[Non-Verifiable]} and assign the factual label as \textbf{[Not Enough Information]}.

\textbf{2. Evidence Selection}

Manually select evidence for the last utterance from Speaker B.

\textbf{3. Claim Verification}

\begin{itemize}
  \item If the utterance is an independent atomic fact, verify it using the selected evidence directly.
  \item If it involves coreference to earlier dialogue, use both the selected evidence and previous dialogue to verify it.
\end{itemize}

Finally, assign the \textbf{Factual Label}:
\begin{itemize}
  \item \textbf{Supports}: The evidence supports the factual claim.
  \item \textbf{Refutes}: The evidence contradicts the factual claim.
  \item \textbf{Not Enough Information}: Evidence is missing or insufficient.
\end{itemize}

\textbf{Note:} If the response is irrelevant to the context, treat it as a standalone factual claim.

\textbf{Summary of Options:}

\textbf{1. Factual Claim}

\textbf{NON-VERIFIABLE}: No verifiable factual info; includes personal opinions or private info.

\textbf{VERIFIABLE}: Contains verifiable factual info checkable via background corpus (e.g., Wikipedia).

\textbf{2. Factual Label}

\textbf{Supports}: Evidence supports the factual claim.

\textbf{Refutes}: Factual claim contradicts the evidence.

\textbf{Not Enough Information}: No or insufficient evidence to verify the claim.
\\
\bottomrule
\end{tabular}
\caption{The instructions for dialogue factual annotation.}
\label{table:factual_annotation_instructions}
\end{table*}

\end{document}